\documentclass[runningheads]{llncs}
\usepackage{graphicx}
\usepackage[normalem]{ulem}

\usepackage{tikz}
\usepackage{comment}
\usepackage{amsmath,amssymb} %
\usepackage{color}
\usepackage{hyperref}
\usepackage[accsupp]{axessibility}  %
\usepackage{microtype}

\usepackage[width=122mm,left=12mm,paperwidth=146mm,height=193mm,top=12mm,paperheight=217mm]{geometry}

\usepackage{adjustbox}[Export]

\usepackage{epsfig}
\usepackage{float}
\usepackage{wrapfig}
\usepackage{algorithm,algorithmicx,algpseudocode}
\usepackage{bm,xspace}
\usepackage{multirow}
\usepackage{balance}
\usepackage{url}
\usepackage{booktabs}
\usepackage{etoolbox,siunitx}
\usepackage{calc}
\usepackage{pifont,hologo}
\usepackage{nicefrac}
\usepackage{amsfonts}
\usepackage{amsmath}

\usepackage[super]{nth}
\usepackage{nicefrac}
\robustify\bfseries

\def\tt{\mathbf{t}}

\DeclareMathSymbol{@}{\mathord}{letters}{"3B}

\def\latex/{\LaTeX}
\def\bibtex/{\hologo{BibTeX}}

\graphicspath{{../images/draft/}{../images/}}

\hypersetup{
    colorlinks=true,
    linkcolor=blue,
    filecolor=magenta,      
    urlcolor=cyan}

\begin{document}
\pagestyle{headings}
\mainmatter
\def\ECCVSubNumber{1424}  %

\newcommand{\modelname}{{ARF}} 

\title{\modelname: Artistic Radiance Fields}

\titlerunning{\modelname: Artistic Radiance Fields}

\author{Kai Zhang$^1$
\quad
Nick Kolkin$^2$
\quad
Sai Bi$^2$
\quad
Fujun Luan$^2$ \\
\quad
Zexiang Xu$^2$
\quad
Eli Shechtman$^2$
\quad
Noah Snavely$^1$
}
\institute{$^1$Cornell University \quad $^2$Adobe Research}
\authorrunning{K.\ Zhang et al.}

\maketitle

\newcommand{\sai}[1]{{\color{magenta}{Sai: #1}}}
\newcommand{\noah}[1]{\noindent {\color{purple} [{\bf Noah:}   {#1}]}}
\newcommand{\kz}[1]{\noindent {\color{red}    {\bf Kai:}     {#1}}}
\newcommand{\fl}[1]{\noindent {\color{orange}   {\bf F:}   {#1}}}

\begin{abstract}

We present a method for transferring the artistic features of an arbitrary style image to a 3D scene. 
Previous methods that perform 3D stylization on point clouds or meshes are sensitive to geometric reconstruction errors for complex real-world scenes. Instead, we propose to stylize the more robust radiance field representation. We find that the commonly used Gram matrix-based loss  tends to produce blurry results without faithful brushstrokes, and introduce a nearest neighbor-based loss that is highly effective at capturing style details while maintaining multi-view consistency. We also propose a novel deferred back-propagation method to enable optimization of memory-intensive radiance fields using style losses defined on full-resolution rendered images. 
Our extensive evaluation demonstrates that our method outperforms baselines by generating artistic appearance that more closely resembles the style image.
Please check our project page for video results and open-source implementations: \url{https://www.cs.cornell.edu/projects/arf/}.

\keywords{Style transfer, neural radiance fields, 3D content creation}

\end{abstract}

\section{Introduction}
Creating artistic images often
requires a significant amount of time and special expertise. 
Extending an artwork to dimensions beyond the 2D image plane, such as time (in the case of animation), or 3D space (in the case of sculptures or virtual environments), introduces new constraints and challenges. 
Hence, the styles employed by artists when moving their work beyond a static 2D canvas 
are constrained by the effort required to create a consistent visual experience. 

\begin{figure}
    \centering
    \adjustimage{width=1.0\textwidth}{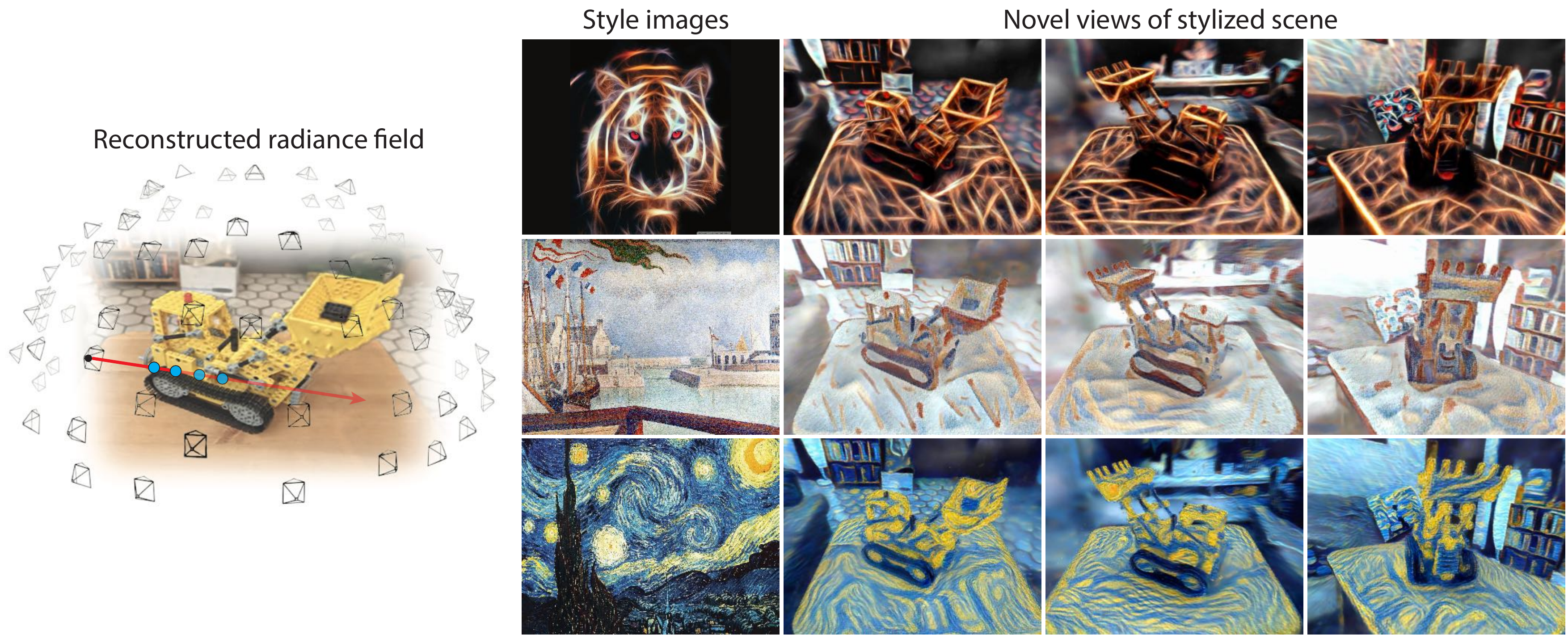}
    \caption{
    We propose \modelname, \ a novel approach for 3D stylization.
    Our approach utilizes a pre-reconstructed radiance field of a real scene (left) and converts it into an artistic radiance field by matching feature activations extracted from an input 2D style image (middle), leading to high-quality stylized novel view synthesis (right). Our approach produces consistent results across viewpoints; please refer to our supplementary video.
    }
    \label{fig:teaser}
\end{figure}

We propose Artistic Radiance Fields (\modelname), a novel approach that can transfer
the artistic features from a single 2D image to 
a full, real-world 3D scene, leading to artistic novel view renderings that are faithful to the style image.
Our method converts a photorealistic radiance field~\cite{mildenhall2020nerf,yu2021plenoxels,TensoRF} reconstructed from multiple images of complex, real-world scenes into a new, stylized radiance field that supports high-quality view-consistent stylized renderings from novel viewpoints,
as shown in Fig.~\ref{fig:teaser}. The quality of these renderings is in contrast to previous 3D stylization works~\cite{huang2021learning,hollein2021stylemesh,mu20213d} that often suffer from geometrically inaccurate reconstructions of point cloud or triangle meshes and the lack of style details.

We formulate the stylization of radiance fields as an optimization problem; we render images of the radiance fields from different viewpoints in a differentiable manner, and minimize 
a content loss between the rendered stylized images and the original captured images, and  also 
a style loss between the rendered images and the style image.
While previous methods~\cite{huang2021learning,hollein2021stylemesh,mu20213d}  apply the commonly-used Gram matrix-based style loss 
for 3D stylization, we observe that such a loss 
leads to averaged-out style details that degrades the quality of the stylized renderings.

This limitation motivates us to apply a novel style loss 
based on Nearest Neighbor Feature Matching (NNFM)
that is better suited to the creation of high-quality 3D artistic radiance fields. 
In particular, for each feature vector in the VGG feature map of a rendered image, we find its nearest neighbor (NN) feature vector 
in the style image's VGG feature map and minimize the distance between the two feature vectors. Unlike a Gram matrix describing global feature statistics across the entire image, NN feature matching focuses on local image descriptions, better capturing distinctive local details.
Coupled with our style loss, we also enforce a VGG feature-based content loss --  that balances stylization and content preservation --  and a simple color transfer technique -- that improves the color match between our final renderings and the input style.

Volumetric radiance field rendering consumes a lot of memory and often can only regress sparsely sampled pixels during  training, and not the full images necessary for computing the VGG features used in many style losses.
We contribute a practical innovation 
that allows us to perform optimization on high-resolution images. 
In particular, we devise a method we call \emph{deferred back-propagation} that enables memory-efficient auto-differentiation of scene parameters with image losses computed on full-resolution images (e.g., VGG-based style losses) by accumulating cached gradients in a patch-wise fashion.

We demonstrate that \modelname \ can robustly transfer detailed artistic features from diverse and challenging 2D style exemplars to a variety of complex 3D scenes, resulting in significantly better visual quality compared to previous methods, which tend to yield over-smoothed and blurry stylized novel views (see Figures~\ref{fig:ff_compare}, \ref{fig:tt_compare}, and \ref{fig:moreresults}). 
In our user studies, our method is also
consistently preferred over baselines.

In summary, our contributions are:
\begin{itemize}
    \item A novel radiance field-based approach for 3D scene stylization that can faithfully transfer detailed style features from a 2D image to a 3D scene and produces consistent stylized novel views of high visual quality.
    \item We find that Nearest Neighbor Feature Matching (NNFM) loss better preserves details in the style images than the Gram-matrix-based loss commonly used in prior 3D stylization works.

    \item A deferred back-propagation method for differentiable volumetric rendering, allowing for computation of losses on full-resolution images while significantly reducing the GPU memory footprint.

\end{itemize}

\section{Related Work}
In this section, we review related work to provide context for our own work.

\medskip
\noindent \textbf{Image style transfer}. 
Since Gatys et al.~\cite{gatys2016image} introduced neural style transfer, significant progress has been made towards artistic stylization~\cite{liao2017visual,li2016combining}, image harmonization~\cite{zhang2020deep,luan2018deep,tsai2017deep}, color matching~\cite{xia2020joint,xia2021real,luan2017deep}, texture synthesis~\cite{risser2017stable,li2017diversified,heitz2021sliced} and beyond~\cite{jing2019neural}. These style transfer approaches leverage features extracted by a pre-trained convolutional neural network (e.g., VGG-19~\cite{simonyan2014very}) and optimize for a set of loss functions (typically a content loss capturing an input photo's features and a style loss matching a target image's feature statistics, e.g., encoded in a Gram matrix) to achieve 
good performance for painterly style transfer. 
Depending on whether the style transfer is achieved via iterative optimization on a single input or with a forward pass from a pre-trained generative model, existing methods can be categorized as optimization-based and feed-forward-based:   

\medskip
\emph{Optimization-based style transfer}. 
Gatys et al.~\cite{gatys2016image} perform style transfer via iterative optimization to minimize content and style losses.
Many follow-up works~\cite{chen2016fast,li2016combining,risser2017stable,gu2018arbitrary,kolkin2019style,mechrez2018contextual,liao2017visual} have investigated alternative style loss formulations to further improve the quality of semantic consistency and high-frequency style details like 
brushstrokes. Unlike neural style transfer methods that encode statistics of style features with a single Gram matrix, Chen and Schmidt \cite{chen2016fast}, CNNMRF~\cite{li2016combining}, Deep Image Analogy~\cite{liao2017visual} and NNST~\cite{kolkin2022neural} propose to search for nearest neighbors and minimize distances between features extracted from corresponding content and style patches in a coarse-to-fine fashion. These methods achieve impressive 2D stylization quality when provided with source and target images that share similar semantics. 
Our approach draws inspiration from this line of work and is the first to introduce nearest neighbor feature matching (NNFM) for 3D stylization.
Our NNFM loss is most similar to that proposed in \cite{kolkin2022neural} for 2D style transfer. However, when stylizing 3D radiance fields, we find that we can achieve the same level of stylistic detail more efficiently by only applying stylization at the final scale (as opposed to coarse-to-fine) and by skipping the style image augmentations (rotation and/or scaling) used in \cite{li2016combining,kolkin2022neural}.

\medskip 
\emph{Feed-forward style transfer}. 
Rather than performing iterative optimization, feed-forward approaches~\cite{huang2017arbitrary,an2021artflow,chiu2020iterative,park2019arbitrary,yao2019attention,sheng2018avatar,li2017universal} train neural networks that can capture 
the style information of the style image and transfer it to the input image using a single forward pass.
While fast, 
these methods often struggle to  faithfully reproduce stylistic feature like colors and brushstrokes, and yield lower visual quality compared to optimization-based techniques. For the sake of creating high-quality artistic radiance fields, we do not pursue this direction as a component of \modelname.

\medskip 
\noindent \textbf{Video style transfer.} Stylizing each video frame separately with a 2D style transfer method often leads to flickering artifacts in the resulting stylized videos. 
Video style transfer~\cite{ruder2018artistic} techniques address this problem by enforcing an additional temporal coherency loss across frames~\cite{chen2017coherent,huang2017real,ruder2018artistic,wang2020consistent}. 
Alternative approaches rely on aligning and fusing style features according to 
their similarity to content features~\cite{deng2020arbitrary,liu2021adaattn} to maintain temporal consistency. Despite sharing the similar challenge of consistency across views, stylizing a 3D scene is a distinct problem from video stylization, because it requires synthesizing novel views while maintaining style consistency, which in turn is best achieved through stylization in 3D rather than 2D image space.

\medskip
\noindent \textbf{3D style transfer.} 
3D style transfer aims to transform the appearance of a 3D scene so that its renderings from different viewpoints match the style of a desired image. Previous approaches represent real world scenes using point clouds~\cite{huang2021learning,mu20213d} or triangle meshes~\cite{yin20213dstylenet,michel2021text2mesh}. For example, Huang et al.~\cite{huang2021learning} and Mu et al.~\cite{mu20213d} use featurized 3D point clouds modulated with the style image, followed by a 2D CNN renderer to produce stylized renderings. Yin et al.~\cite{yin20213dstylenet}  create novel geometric and texture variations of 3D meshes by transferring the shape and texture style from one textured mesh to another. The performance of such methods is limited by the quality of the geometric reconstructions, which oftentimes contain noticeable artifacts for complex real-world scenes.  
In contrast, we perform style transfer on radiance fields~\cite{mildenhall2020nerf,liu2020neural,yu2021pixelnerf,zhang2020nerf++,chen2021mvsnerf} which have been shown to 
more faithfully reproduce the appearance of real world scenes. A work closely relevant to ours is that of Chiang et al.~\cite{chiang2022stylizing}, who apply neural radiance fields for scene representation and rely on pre-trained style hypernetworks for appearance stylization. However, their method produces over-smoothed and blurry stylization results, and cannot capture the detailed structures of the style image such as brushstrokes, due to the limitation of pre-trained feed-forward models. We show that our approach can more faithfully capture distinctive details in the style exemplar while preserving recognizable scene content.

\section{Background of Radiance Fields}
NeRF \cite{mildenhall2020nerf} proposes neural radiance fields to model and reconstruct real scenes, achieving photo-realistic novel view synthesis results.
In general, the radiance field representation can be seen as a 5D function that maps any 3D location $\mathbf{x}$ and viewing direction $\mathbf{d}$ to volume density $\sigma$ and RGB color $\mathbf{c}$:
\begin{equation}
    \sigma, \mathbf{c} = \textsc{RadianceField}(\mathbf{x}, \mathbf{d}).
\end{equation}
This representation can be rendered from any viewpoint via differentiable volume rendering, and hence can be optimized to fit a collection of input photos captured from multiple views, and then later used for synthesizing photo-realistic novel views. 
We move beyond photo-realism and add an artistic feel to the radiance field by stylizing it using an exemplar style image, such as a painting or sketch.

\section{Stylizing Radiance Fields}

\begin{figure}[!t]
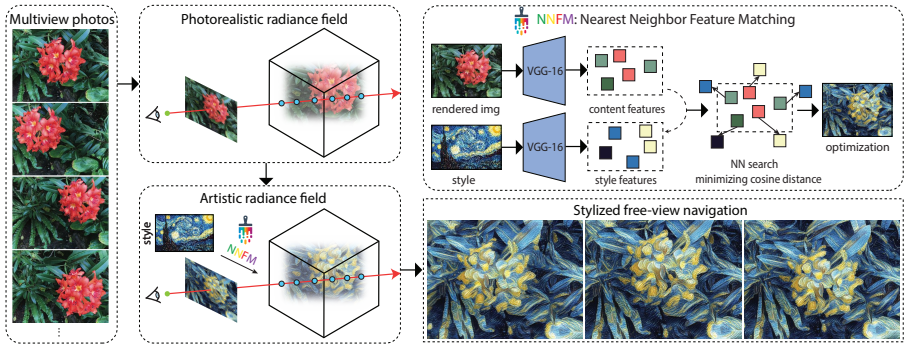

    \centering
    \adjustimage{width=1.0\textwidth}{figure/fig2_v4}
    \caption{Overview of our method. We first reconstruct a photo-realistic radiance field from multiple photos. We then stylize this reconstruction using an exemplar style image through the use of a nearest neighbor feature matching (NNFM) style loss.
    Once this stylization is done, we can obtain consistent free-viewpoint stylized renderings. We invite readers to watch the supplemental videos to better appreciate our results.
    }
    \label{fig:method_overview}
\end{figure}

In this section, we describe our radiance fields stylization technique in detail. Given a photo-realistic radiance field reconstructed from photos of a real scene, our approach transforms it into an artistic style by stylizing the 3D scene appearance with a 2D style image. We achieve this by fine-tuning the radiance field using a novel nearest neighbor feature matching style loss (Sec.~\ref{sec:styleloss}) that can transfer detailed local style structures. We also introduce a deferred back-propagation technique that enables radiance field optimization with full-resolution images (Sec.~\ref{sec:deferredbg}) in the face of limited GPU memory. 
We apply a view-consistent color transfer technique to further enhance our final visual quality (Sec.~\ref{sec:colortransfer}).

\subsection{Style transfer losses}
\label{sec:styleloss}
Artwork often features unique visual details; for instance, the Van Gogh's \emph{The Starry Night} is characterized by 
long and curvy brushstrokes. 
In general, neural features produced by pre-trained neural networks (like VGG) can effectively capture such 
details and have been successfully used for 2D style transfer~\cite{gatys2016image}.
However, it is challenging to transfer such rich visual details to 3D scenes using prior VGG-based style losses, since the style information measured by such losses are generally based on global statistics that do not necessarily capture the local details well in a view-consistent way.

To address this, we propose to use the \textit{\textbf{N}earest \textbf{N}eighbor \textbf{F}eature \textbf{M}atching} (NNFM) loss to transfer complex high-frequency visual details from a 2D style image to a 3D scene (parameterized by a radiance field), consistently across multiple viewpoints.
In particular, let $\boldsymbol{I}_{\textrm{style}}$ denote the style image, and $\boldsymbol{I}_{\textrm{render}}$ denote an image rendered from the radiance field at a selected viewpoint. We extract the VGG feature map $\boldsymbol{F}_{\textrm{style}}$ and ${\boldsymbol{F}_{\textrm{render}}}$
for $\boldsymbol{I}_{\textrm{style}}$ and ${\boldsymbol{I}}_{\textrm{render}}$, respectively.
Let ${\boldsymbol{F}}_{\textrm{render}}(i, j)$ denote the feature vector at pixel location $(i, j)$ of the feature map ${\boldsymbol{F}}_{\textrm{render}}$.
Our NNFM loss can be written as: 
\begin{align}
\ell_{\textrm{nnfm}}({\boldsymbol{F}_{\textrm{render}}}, \boldsymbol{F}_{\textrm{style}})&=\frac{1}{N}\sum_{i, j} \min_{i', j'} D\big({\boldsymbol{F}_{\textrm{render}}}(i, j),\boldsymbol{F}_{\textrm{style}}(i',j')\big), \label{eq:nnfm}
\end{align}
where $N$ is the number of pixels in ${\boldsymbol{F}_{\textrm{render}}}$, and $D(\boldsymbol{v}_1, \boldsymbol{v}_2)$ computes the cosine distance between two vectors $\boldsymbol{v}_1, \boldsymbol{v}_2$: 
 \begin{align}
      D(\boldsymbol{v}_1, \boldsymbol{v}_2)=1-\boldsymbol{v}_1^T\boldsymbol{v}_2/\sqrt{\boldsymbol{v}_1^T\boldsymbol{v}_1 \boldsymbol{v}_2^T\boldsymbol{v}_2}. \label{eq:cosine_dist}
 \end{align}
In short, for each feature in ${\boldsymbol{F}_{\textrm{render}}}$, we minimize its cosine distance (Eq.~(\ref{eq:cosine_dist})) to its nearest neighbor in the style image's VGG feature space ($\boldsymbol{F}_{\textrm{style}}$).

Note that our loss does not rely on global statistics. This grants more flexibility to the optimization process, which can focus on adjusting the local scene appearance to perceptually match the style image in a given image rendered from a given training viewpoint.

\subsubsection{Controlling stylization strength.}
Using our NNFM loss alone can sometimes lead to overly strong stylization, making the content harder to recognize. To address this issue, we add an additional content-preserving loss penalizing the $\ell_2$ difference between the feature maps of rendered and content images: 
\begin{align}
    \ell=\ell_{\textrm{nnfm}}({\boldsymbol{F}_{\textrm{render}}}, \boldsymbol{F}_{\textrm{style}})+\lambda \cdot \ell_2({\boldsymbol{F}_{\textrm{render}}}, \boldsymbol{F}_{\textrm{content}}),
    \label{eqn:finalloss}
\end{align}
where $\lambda$ is a weight controlling stylization strength: a larger $\lambda$  preserves more content, while a smaller $\lambda$ leads to stronger stylization. Note that  ${\boldsymbol{F}_{\textrm{render}}}, \boldsymbol{F}_{\textrm{style}}, \boldsymbol{F}_{\textrm{content}}$ are extracted by exactly the same feature extractors. (See Sec.~\ref{sec:impl_details} for details)

\subsection{Deferred back-propagation}

\begin{figure}[!t]
\centering
\includegraphics[width=1.0\textwidth]{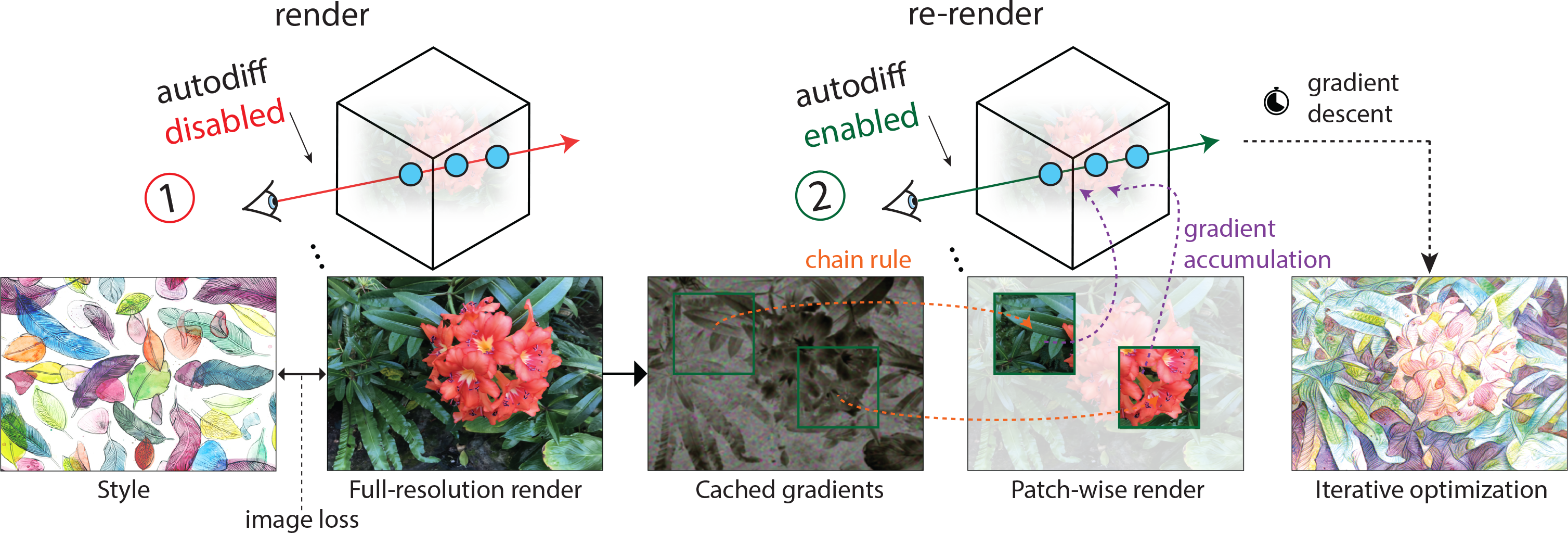}
\caption{Illustration of deferred back-propagation. We first disable auto-differentiation to render a full-resolution image; then we compute the image loss (e.g., a style loss defined by NNFM or by a Gram matrix), and cache its gradients with respect to pixel values of the full-resolution image. Next, we back-propagate the cached gradients to scene parameters and accumulate in a patch-wise manner; for each patch, we re-render it with enabled auto-differentiation, and apply the chain rule to back-propagate the corresponding cached patch gradients to scene parameters for accumulation. This way, we correctly compute the gradients of a loss imposed on the full-resolution rendered image with respect to the scene parameters, with the same GPU memory footprint of rendering a single small patch differentiably. 
}\label{fig:deferred}
\end{figure}

\label{sec:deferredbg}
We stylize a radiance field  by minimizing our loss (Eqn.~\ref{eqn:finalloss}) imposed on images rendered using differentiable volume rendering. %
However, such rendering is very memory-inefficient in practice, because the color of each pixel is composited from a large number of samples along the ray. As a result, rather than rendering a full-resolution image at each optimization iteration, many methods randomly sample a sparse set of pixels for rendering. 
While such sparse pixel sampling is a reasonable strategy when minimizing a loss computed independently per-pixel, such as an $\ell_1/\ell_2$ loss, it does not work for complex CNN-based losses, such as our NNFM loss or a Gram-matrix style loss, which require full-resolution images.

We propose a simple technique termed \emph{deferred back-propagation} that can directly optimize on full-resolution images, allowing for more sophisticated and powerful image losses to be used in practice with radiance fields representation. 
As shown by Fig.~\ref{fig:deferred}, 
we first render a full-resolution image with auto-differentiation disabled; then we compute the image loss and its gradient with respect to the rendered image's pixel colors, which produces a cached gradient image; finally, in a \textit{patch-wise} manner, we re-render the pixel colors with \emph{enabled} auto-differentiation, and back-propagate the cached gradients to the scene parameters for accumulation. In this way, gradient back-propagation is deferred from the full-resolution image rendering stage to the patch-wise re-rendering stage, reducing the GPU memory cost from that of rendering a full-resolution image to that of rendering a small patch.
In our work we apply this technique for the stylization task by optimizing our style loss, and also by optimizing the standard Gram loss for comparison (see Fig.~\ref{fig:abl}).

\subsection{View-consistent color transfer}
\label{sec:colortransfer}
While our style and content losses can perceptually transfer styles and preserve the original content, we find they can lead to color mismatches between rendered images and the style image. 
We devise a simple technique to address this issue which leads to much better stylization quality (see Fig.~\ref{fig:abl}).
We first recolor the training views via color transfer from the style image.
These recolored images are used to pre-optimize our artistic radiance field as initialization for our stylization optimization based on Eq.~\ref{eqn:finalloss}. These color transferred images are also used for our content preservation loss. Additionally, after the 3D stylization process, we 
again perform a color transfer to images rendered to the training viewpoints, and apply the same color transformation directly to the color values produced from rendering the radiance fields.

As to the color transfer method, we adopt a simple linear transformation of colors in RGB space, parameters of which are estimated by matching color statistics  of an image set to those of an image.
Specifically, let $\big\{\boldsymbol{c}_i\big\}_{i=1}^m$ be the set of all pixel colors in an image set to be recolored, and let $\big\{\boldsymbol{s}_i\big\}_{i=1}^n$ be the set of all pixel colors of the style image; we analytically solve for a linear transformation $\boldsymbol{A}$ such that $\mathrm{E}\big[\boldsymbol{A} \boldsymbol{c}]=\mathrm{E}\big[\boldsymbol{s}\big]$ and %
$\mathrm{Cov}\big[\boldsymbol{Ac}\big]=\mathrm{Cov}\big[\boldsymbol{s}\big]$, i.e., the mean and covariance of the color-transformed image set match those of the style image.

\subsection{Implementation details} \label{sec:impl_details}
\begin{figure*}[!t]
\centering 
\includegraphics[width=0.98\textwidth]{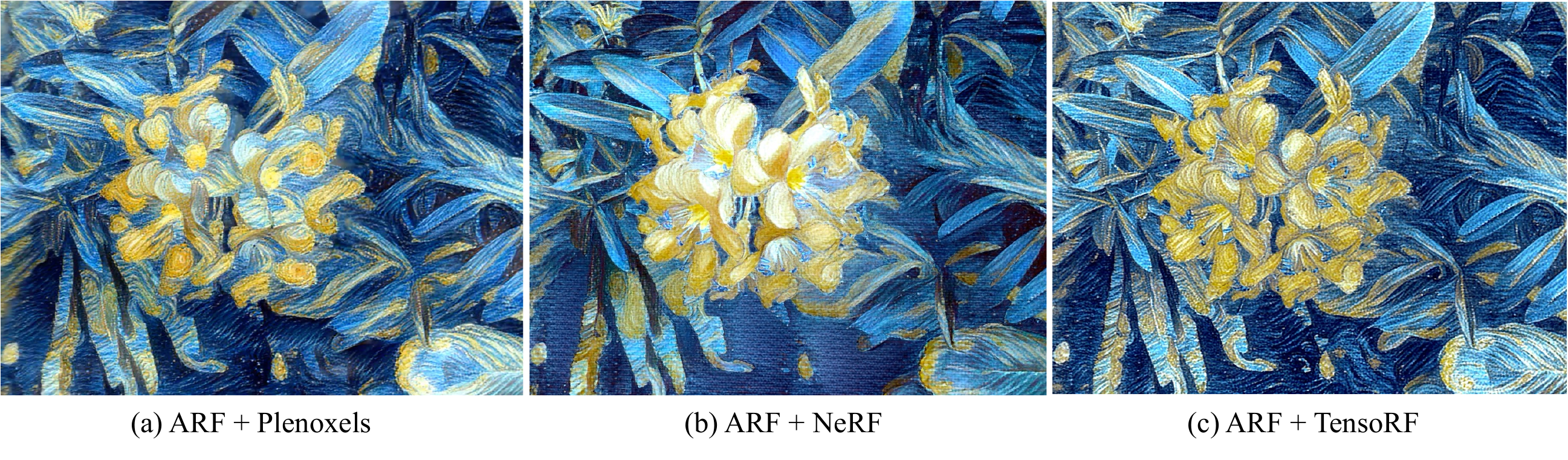}
\caption{Applicability across different radiance field representations.  Our ARF method is applicable to a variety of radiance fields representations, including Plenoxels~\cite{yu2021plenoctrees}, NeRF~\cite{mildenhall2020nerf} and TensoRF~\cite{TensoRF}, in each case producing high-quality 3D stylization results. 
}\label{fig:nerf_variants}
\end{figure*}

To represent a radiance field, our work primarily uses the recently-proposed Plenoxels~\cite{yu2021plenoctrees} for its fast reconstruction and rendering speed. 
However, our framework is agnostic to the radiance field representation. To demonstrate this, we apply our proposed techniques to stylize NeRF~\cite{mildenhall2020nerf} and TensoRF~\cite{yu2021plenoctrees}, and in each case achieve high visual quality with faithful style transfer, as shown in Fig.~\ref{fig:nerf_variants}. 

During stylization, we fix the density component of the initial photorealistic radiance field, and only optimize the appearance component when converting to an artistic radiance field. We also discard the view-dependent appearance modelling,\footnote{In Plenoxels, radiance fields are represented as a mixture of spherical harmonics at each point. To discard view-dependence, we simply move all  spherical harmonics components except the first one. For TensoRF, we zero out the view directions when inputting them to the MLP. }
To extract feature maps  ${\boldsymbol{F}_{\textrm{render}}}, \boldsymbol{F}_{\textrm{style}}, \boldsymbol{F}_{\textrm{content}}$ in Eq.~\ref{eqn:finalloss}, we use a pretrained VGG-16 network that consists of 5 layer blocks: \texttt{conv1}, \texttt{conv2}, \texttt{conv3}, \texttt{conv4}, \texttt{conv5}.\footnote{In VGG-16, each layer block begins with a max-pooling layer that downsamples  the feature map by 2. Inside each layer block, feature maps are of the same spatial resolution and hence can be concatenated to form a single feature map for this block.} We use the \texttt{conv3} block as the feature extractor, because we empirically find that it captures style details better than the other blocks, as shown in Fig.~\ref{fig:ablation_vgg}. We set the content-preserving weight $\lambda=0.001$ in Eqn.~\ref{eqn:finalloss} for all forward-facing captures, and $\lambda=0.005$ for all 360$^\circ$ captures. At each stylization iteration, we render an image for computing losses from a viewpoint randomly selected out of all the training viewpoints used for reconstructing photo-realistic radiance fields. We perform the stylization optimization for 10 epochs with learning rate exponentially decayed from 1e-1 to 1e-2.

\section{Experiments}
We evaluate our method by performing both quantitative and qualitative comparisons to baseline methods.
We show stylization results for various real world scenes guided by different style images. The experimental results show that our method significantly outperforms baseline methods in terms of generating stylized renderings that are more faithful to the input style image, while maintaining the recognizable semantic and geometric features of the original scene. We invite readers to watch our supplemental videos for better assessment of 3D stylization quality. 

\medskip

\noindent\textbf{Datasets.} We conduct extensive experiments on multiple real-world scenes 
including four forward-facing captures: \emph{Flower, Orchids, Horns, Trex}, from~\cite{mildenhall2020nerf}, and seven 360$^\circ$ captures: \emph{Family, Horse, Playground, Truck, M60,  Train} from the Tanks and Temples dataset~\cite{Knapitsch2017}, as well as the \emph{Real Lego}
dataset from~\cite{yu2021plenoxels}. All scenes contain complex structures and intricate details
that are difficult to reconstruct with previous triangle mesh or point cloud-based methods. 
We also experiment with a diverse set of style images 
including a neon tiger, Van Gogh's \emph{The Starry Night}, sketches, etc., to test our method's ability to handle a diverse range of style exemplars.

\medskip
\noindent\textbf{Baselines. } We compare our method to state-of-the-art methods~\cite{huang2021learning,chiang2022stylizing} for 3D style transfer quality. 
Specifically, Huang et al.~\cite{huang2021learning} adopt point clouds featurized by VGG features averaged across views as a scene representation, and transform the pointwise features by modulating them with the encoding vector of a style image for stylization.
Chiang et al.~\cite{chiang2022stylizing} use implict MLPs as in NeRF++~\cite{zhang2020nerf++} to reconstruct a radiance field for a scene, then update the weights of the radiance prediction branch 
using a hypernetwork that takes a style image as input.
For both methods, we use their released code and pre-trained models.
We chose not to compare 
to off-the-shelf video stylization methods, 
because prior work has demonstrated that they are less competitive compared to 3D style transfer approaches~\cite{huang2021learning,chiang2022stylizing}.

\begin{figure*}[htp]
\centering 
\includegraphics[width=0.99\textwidth]{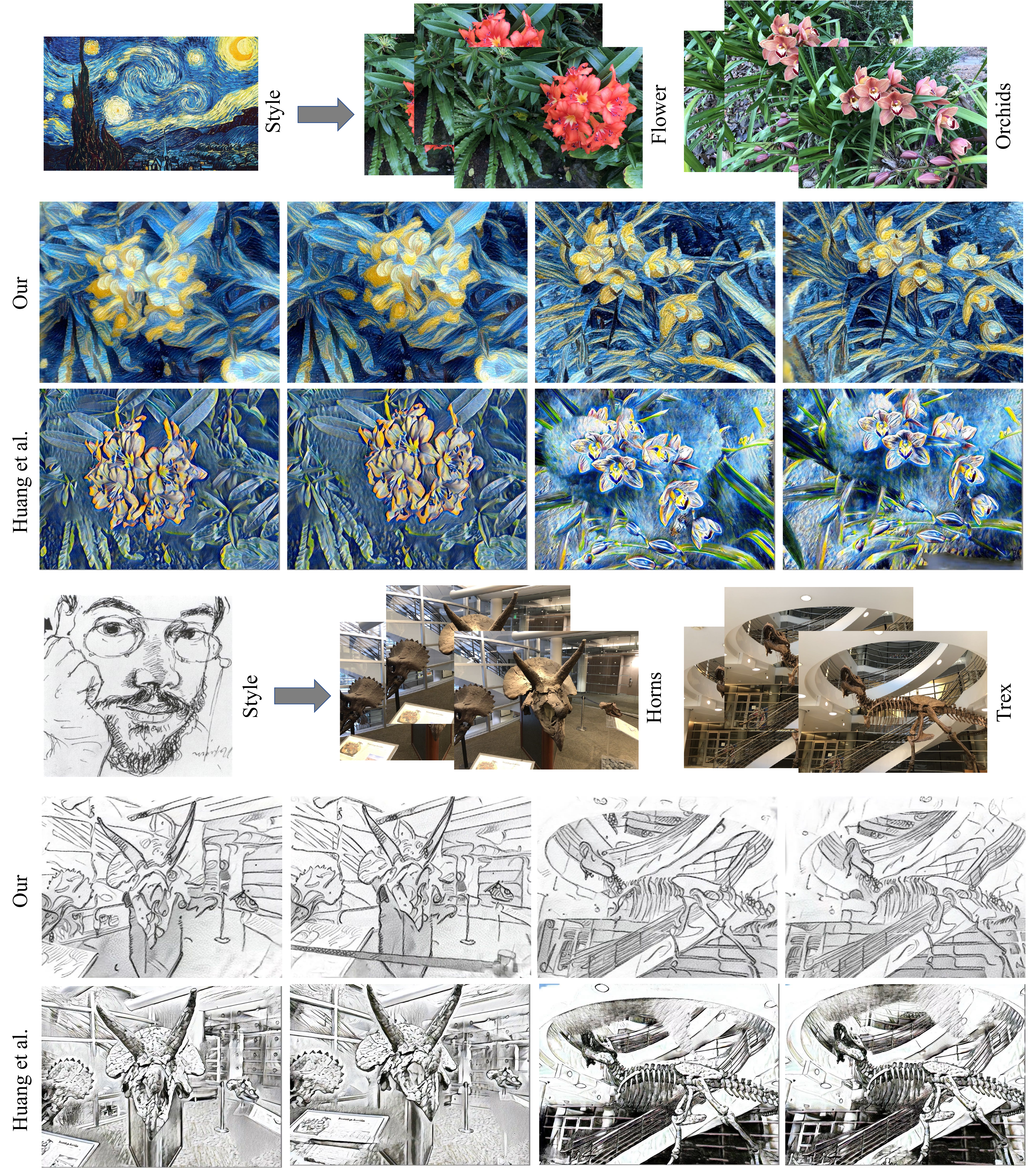}%
\caption{Comparison with the baseline method Huang et al.~\cite{huang2021learning} on real-world forward-facing data. Our stylized novel views contain significantly more faithful style details. Additionally, the method of Huang et al.~\cite{huang2021learning} requires reconstructing meshes from images, an error-prone process. Such errors can impact the quality of stylized novel views, as can be seen in the leaves in the \emph{Flower} and \emph{Orchids} scenes, and in ceiling of \emph{Trex}. In contrast, our method, based on radiance fields, exhibits many fewer geometric artifacts. We invite readers to watch our supplemental videos for better comparisons. 
}
\label{fig:ff_compare}
\end{figure*}

\begin{figure*}[htp]
\centering 
\includegraphics[width=1.0\textwidth]{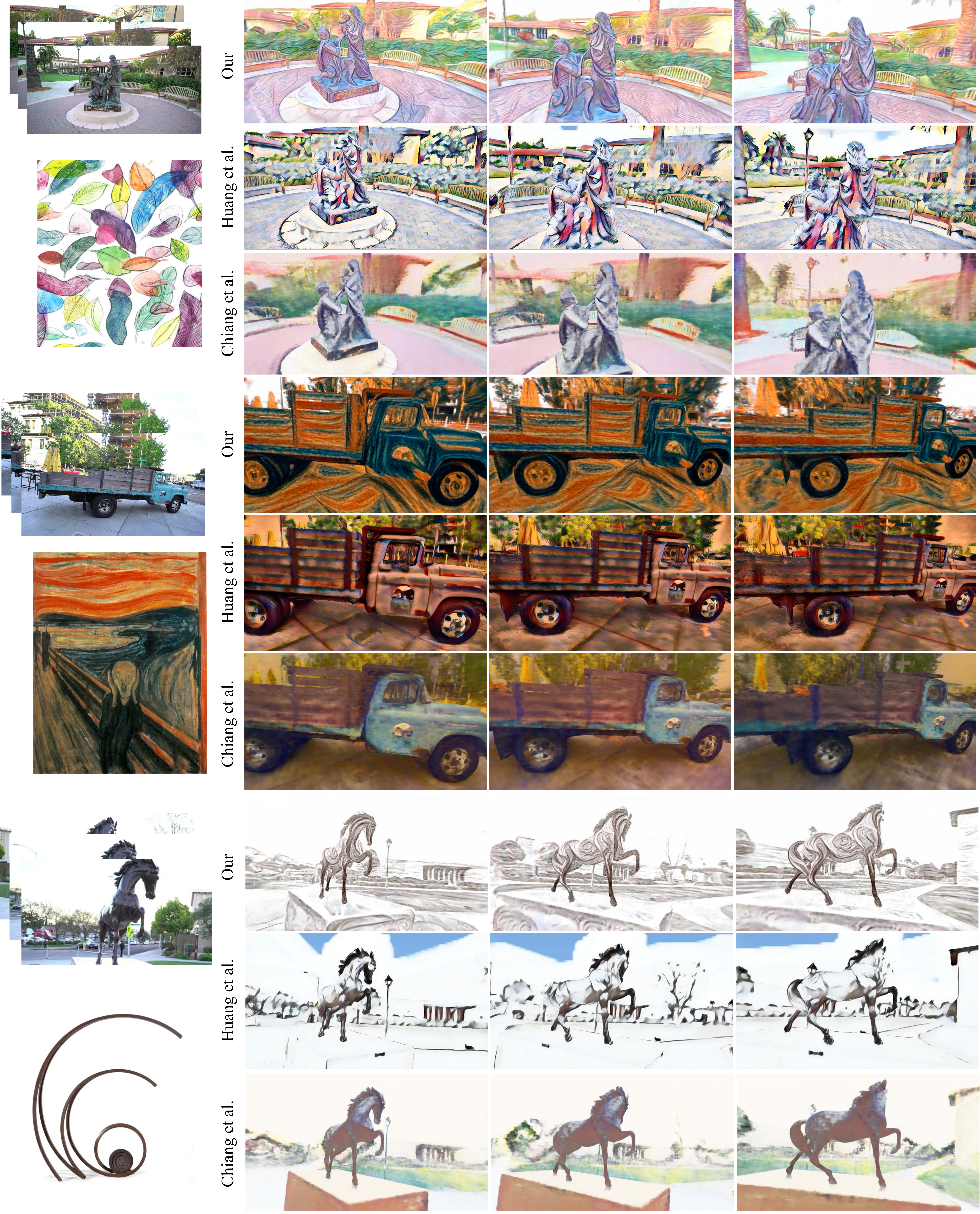}%
\caption{Comparison with the baseline methods Huang et al.~\cite{huang2021learning} and Chiang et al.~\cite{chiang2022stylizing} on real-world Tanks and Temples data. Our results  match both the colors and details of the style image most faithfully, while preserving sharp and recognizable content. 
We invite readers to watch our supplemental videos for better comparisons.}
\label{fig:tt_compare}
\end{figure*}

\medskip
\noindent\textbf{Qualitative comparisons. } 
We show visual comparisons between methods in Fig.~\ref{fig:ff_compare} (forward-facing captures)
and Fig.~\ref{fig:tt_compare} (360$^\circ$ captures). 
Visually, we see that our results 
exhibit a better style match to the exemplar image compared to the baselines.
For instance, in the \emph{Flower} scene in Fig.~\ref{fig:ff_compare}, 
our method faithfully captures both the color tone and the brushstrokes of \emph{The Starry Night}, while the baseline method of Huang et al.~\cite{huang2021learning} generates
over-smoothed results without less detailed structures. Moreover, Huang et al.~also fails to recover complex geometric structures such as leaves of the plants due to inaccuracies in the reconstructed point cloud. 
In comparison, our method effectively reconstructs and preserves the geometric and semantic content of the original scene, thanks to the more robust radiance fields representation.

Chiang et al.~\cite{chiang2022stylizing} only transfers the overall 
color tone of the style image 
to the scene and fails to recover the rich details that our method does. For example, in the \emph{Family} statue scene in Fig.~\ref{fig:tt_compare}, our method captures the subtle textural details of the watercolor feather style image, and reproduces them in the 
stylized renderings.  In contrast, the method of Chiang et al.~\cite{chiang2022stylizing} generates blurry results with no 
such intricate structures, because their hypernetwork is trained on a fixed dataset 
of style images and often fails to capture the details of an unseen style input. Our method benefits from both the optimization-based framework 
as well as our NNFM style loss, which greatly boost the 3D stylization performance. 

\begin{figure*}[htp]
\centering 
\includegraphics[width=1.0\textwidth]{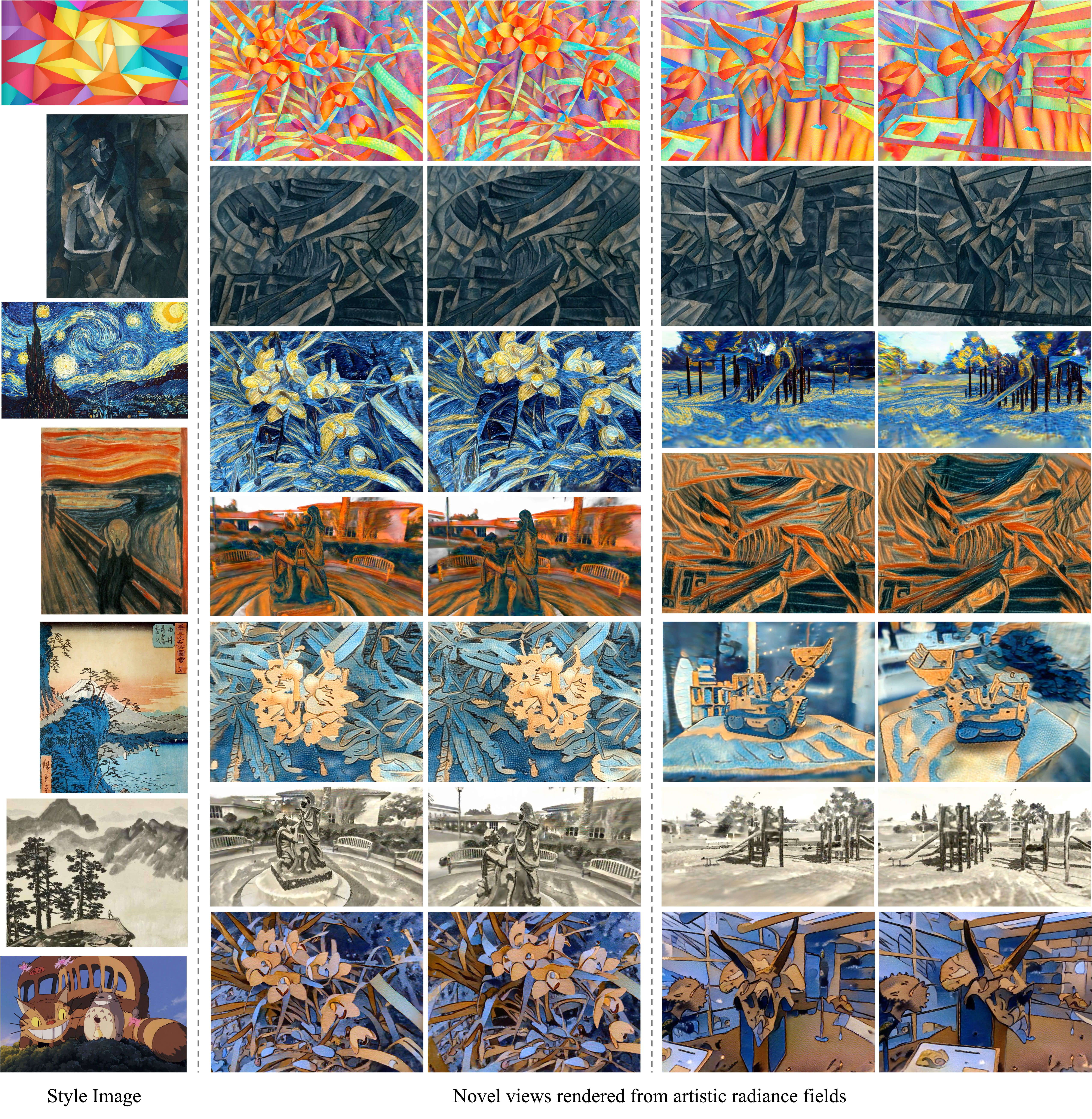}%
\caption{More artistic radiance fields for scenes shown in this paper. The leftmost image in each row is the style image, and the rest are stylized novel views rendered from corresponding artistic radiance fields (two novel views are shown for each artistic radiance field).  Our method can generate compelling results for a wide range of (real world scene, style image) pairs. We refer the readers to our supplemental videos for more visualizations. 
}
\label{fig:moreresults}
\end{figure*}

We show additional results from our method  in Fig.~\ref{fig:moreresults}. 
Our method is robust to different scenes with varying levels of complexity 
and also generates consistently superior results under a variety of styles.  
We refer the readers to the supplementary videos for more visual comparisons 
and results.

\medskip
\noindent\textbf{User study.} We also perform a user study to compare 
our methods to baseline methods. A user is presented with a sequence of stylization results, where for each result the user is shown a style image, a video of the original scene, and two corresponding stylized 
videos produced with our method and a baseline method. The user is then asked to select the result that better matches the style of the given style image. 
In total, we collect ratings covering 25 randomly selected (scene, style) pairs. We divided the questions into 5 batches, each with 5 questions, and asked a large group of users to rate a randomly selected batch. We collected an average of $\sim$12 ratings for each individual pair. 
We found that users prefer our method over the baseline Huang et al.~\cite{huang2021learning} $86.8\%$ of the time, and over the baseline Chiang et al.~\cite{chiang2022stylizing} $94.1\%$ of the time. These results show a  clear preference for our method.

\begin{figure*}[htp]
\centering 
\includegraphics[width=0.98\textwidth]{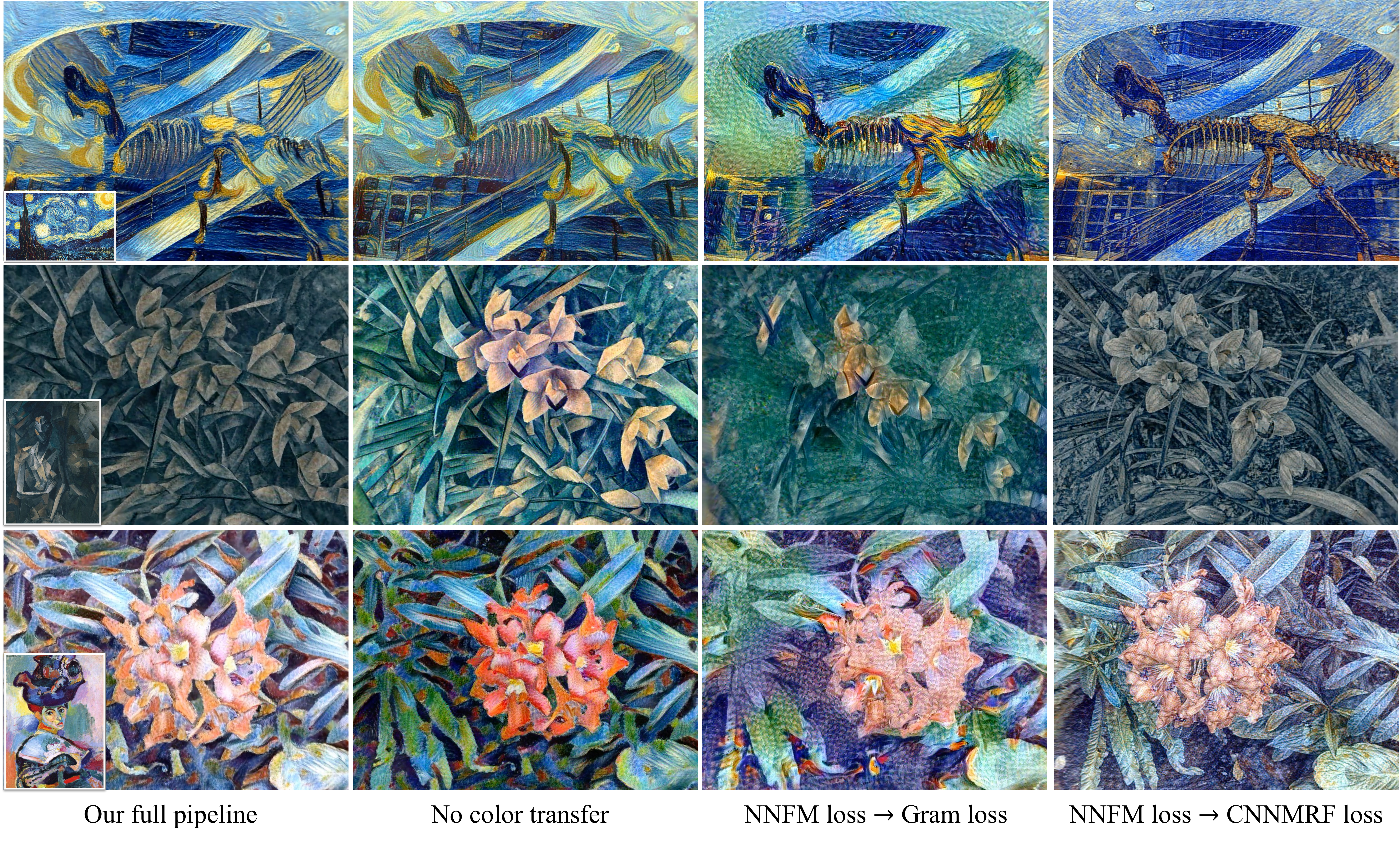}
\caption{Ablation studies of color transfer and NNFM loss. Without color transfer, there is a noticeable color mismatch between the synthesized views and the style image (shown as insets in the first column). Replacing the NNFM loss with the commonly-used Gram loss~\cite{gatys2016image} leads to less compelling results with many more artifacts. Our NNFM loss also generates more faithful 3D stylization results than the prior CNNMRF loss~\cite{li2016combining}. %
}
\label{fig:abl}
\end{figure*}

\begin{figure*}[!t]
\centering 
\includegraphics[width=1.0\textwidth]{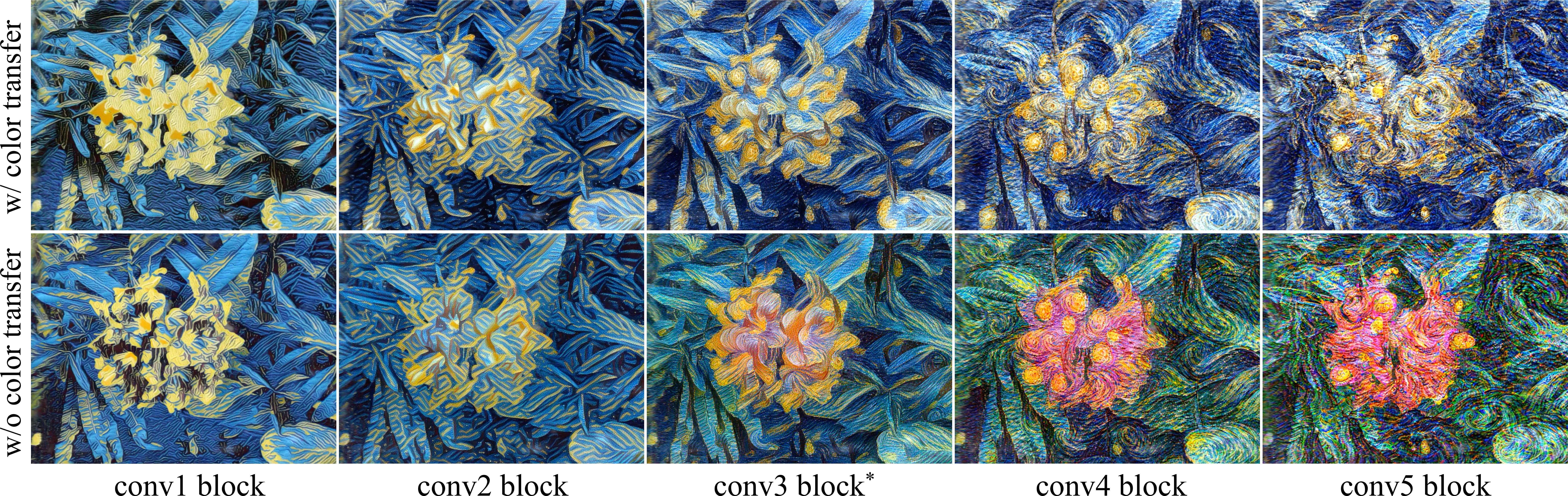}
\caption{Ablation studies of extrating features for losses in Eq.~\ref{eqn:finalloss} using different layer blocks of the VGG-16 network.   We find that conv3 layer block generates more visually pleasing results than other layers. }
\label{fig:ablation_vgg}
\end{figure*}

\medskip
\noindent\textbf{Ablations. } 
We perform ablation studies to justify our design choices. 
We first compare our NNFM loss to the prior Gram-based and CNNMRF losses. 
As we can see from Fig.~\ref{fig:abl}, our NNFM loss generates significantly better 
results and more faithfully preserves the style details of the example images 
compared to the other two losses. In Fig.~\ref{fig:abl}, we also validate the necessity of the color transfer stage. Without color transfer, the generated results tend to 
have different color tones from the style images, leading to a degraded style match. Our color transfer method effectively addresses this issue.
Finally, we perform an ablation of using the feature map at different layers of the VGG-16 network for computing NNFM loss in Fig.~\ref{fig:ablation_vgg}. We find that our choice of the \texttt{conv3} layer block preserves stylistic details better than other layers.

\subsubsection{Limitations.} Our method has a few limitations. First, geometric artifacts, e.g., floaters, in the radiance fields can cause artifacts in both the photorealistic and our stylized renderings. Such floaters can be removed by adding additional regularizers on volume density to the loss during optimization~\cite{barron2022mipnerf360,Niemeyer2021Regnerf}. Second, although our artistic radiance fields can be rendered in real time once optimized, a relatively time-consuming optimization procedure 
is still required for every style image ($\sim$3 mins for forward-facing captures, and $\sim$20 mins for 360 captures on a single NVIDIA RTX 3090 GPU). 
Third, our reconstructed artistic radiance fields do not support manual editing. Enabling artists to interactively edit them is highly desirable for the sake of facilitating creativity.

\section{Conclusion}
We have presented a method to reconstruct artistic radiance fields from photorealistic radiance fields given user-specified style exemplars. The reconstructed artistic radiance fields can then be used to render 
high-quality stylized novel views that faithfully mimic the input style image in terms of 
color tone and 
style details like brushstrokes, enabling an immersive experience of an artistic 3D scene. 
Key to our method's success is the proposed coupling of the nearest neighbor featuring matching loss and view-consistent color transfer, rather than the commonly-used Gram loss. We demonstrate that our method achieves superior 3D stylization quality over baselines through evaluations across various 3D scenes and 2D styles.

\subsubsection{Acknowledgements.} We would like to thank Adobe artist Daichi Ito for helpful discussions about 3D artistic styles.

\bibliographystyle{splncs04}
\bibliography{paper}

\section*{Appendix}
\subsubsection{Derivation of the linear color transformation $\boldsymbol{A}$ in Section 4.3}
\begin{align}
    \boldsymbol{Ac}=\boldsymbol{U_s}\boldsymbol{\Lambda_s}^{\frac{1}{2}}\boldsymbol{U_s}^T\boldsymbol{U_c} \boldsymbol{\Lambda_c}^{-\frac{1}{2}}\boldsymbol{U_c}^T\big(\boldsymbol{c}-\mathrm{E}\big[\boldsymbol{c}\big]\big)+\mathrm{E}\big[\boldsymbol{s}\big],
\end{align}
where $\boldsymbol{U_s},\boldsymbol{\Lambda_s},\boldsymbol{U_c},\boldsymbol{\Lambda_c}$ are obtained via the following eigen-decompositions of covariance matrices $\mathrm{Cov}\big[\boldsymbol{c}\big],\mathrm{Cov}\big[\boldsymbol{s}\big]$:
\begin{align}
       \mathrm{Cov}\big[\boldsymbol{c}\big]&=\boldsymbol{U_c}\boldsymbol{\Lambda_c}\boldsymbol{U_c}^T\\ 
        \mathrm{Cov}\big[\boldsymbol{s}\big]&=\boldsymbol{U_s}\boldsymbol{\Lambda_s}\boldsymbol{U_s}^T.
\end{align}

\noindent \emph{Proof} of $\mathrm{E}\big[\boldsymbol{Ac}\big]=\mathrm{E}\big[\boldsymbol{s}\big]$:
\begin{align}
\mathrm{E}\big[\boldsymbol{Ac}\big]&=\boldsymbol{U_s}\boldsymbol{\Lambda_s}^{\frac{1}{2}}\boldsymbol{U_s}^T\boldsymbol{U_c} \boldsymbol{\Lambda_c}^{-\frac{1}{2}}\boldsymbol{U_c}^T\big(\mathrm{E}\big[\boldsymbol{c}\big]-\mathrm{E}\big[\boldsymbol{c}\big]\big)+\mathrm{E}\big[\boldsymbol{s}\big] \\
&=\boldsymbol{0}+\mathrm{E}\big[\boldsymbol{s}\big]=\mathrm{E}\big[\boldsymbol{s}\big].
\end{align}

\noindent \emph{Proof} of $\mathrm{Cov}\big[\boldsymbol{Ac}\big]=\mathrm{Cov}\big[\boldsymbol{s}\big]$:
\begin{align}
\mathrm{Cov}\big[\boldsymbol{Ac}\big]&=\mathrm{Cov}\big[\boldsymbol{U_s}\boldsymbol{\Lambda_s}^{\frac{1}{2}}\boldsymbol{U_s}^T\boldsymbol{U_c} \boldsymbol{\Lambda_c}^{-\frac{1}{2}}\boldsymbol{U_c}^T\boldsymbol{c}\big]\\
&=\boldsymbol{U_s}\boldsymbol{\Lambda_s}^{\frac{1}{2}}\boldsymbol{U_s}^T\cdot      \mathrm{Cov}\big[\boldsymbol{U_c} \boldsymbol{\Lambda_c}^{-\frac{1}{2}}\boldsymbol{U_c}^T \boldsymbol{c}\big]\cdot (\boldsymbol{U_s}\boldsymbol{\Lambda_s}^{\frac{1}{2}}\boldsymbol{U_s}^T)^T\\
&=\boldsymbol{U_s}\boldsymbol{\Lambda_s}^{\frac{1}{2}}\boldsymbol{U_s}^T\cdot
\boldsymbol{U_c} \boldsymbol{\Lambda_c}^{-\frac{1}{2}}\boldsymbol{U_c}^T\cdot  \mathrm{Cov}\big[\boldsymbol{c}\big] \cdot (\boldsymbol{U_c} \boldsymbol{\Lambda_c}^{-\frac{1}{2}}\boldsymbol{U_c}^T)^T
\cdot (\boldsymbol{U_s}\boldsymbol{\Lambda_s}^{\frac{1}{2}}\boldsymbol{U_s}^T)^T\\
&=\boldsymbol{U_s}\boldsymbol{\Lambda_s}^{\frac{1}{2}}\boldsymbol{U_s}^T\cdot
\boldsymbol{I}
\cdot (\boldsymbol{U_s}\boldsymbol{\Lambda_s}^{\frac{1}{2}}\boldsymbol{U_s}^T)^T\\
&=\mathrm{Cov}\big[\boldsymbol{s}\big].
\end{align}

\begin{figure*}[!t]
\centering 
\includegraphics[width=1.\textwidth]{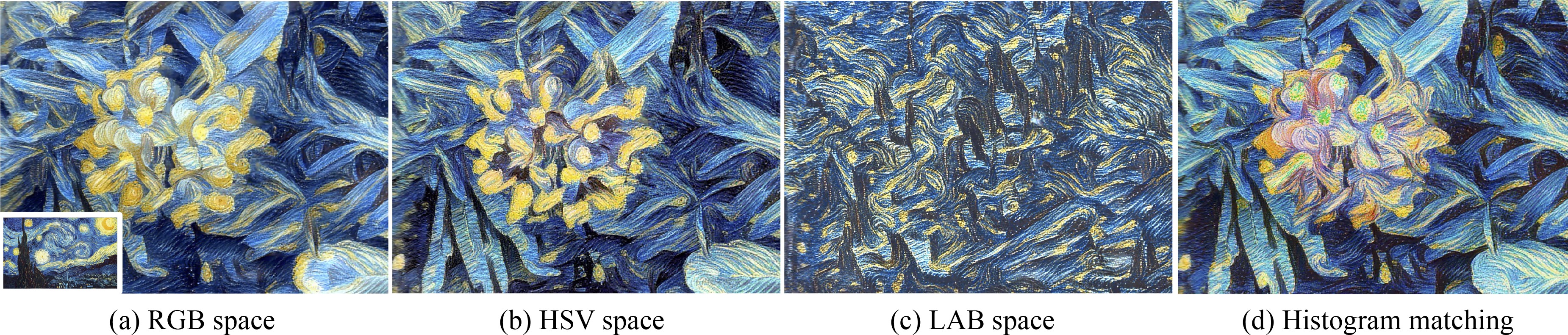}
\caption{Ablation studies of color transfer methods. We execute the color transfer algorithm described in Sec. 4.3 in RGB color space (a); this tends to generate better results than running it in HSV (b) and LAB (c) color spaces, and replacing it with histogram matching (d).  
}\label{fig:colorspaces}
\end{figure*}

\end{document}